\documentclass[letters]{fcs}
\usepackage{layout}
\usepackage{subfig}
\usepackage{hyperref}

\usepackage{algorithm}
\usepackage{algorithmic}
\usepackage{amsmath}
\usepackage{amssymb}
\usepackage{bm}
\usepackage{multirow}
\usepackage{subcaption}
\usepackage{wrapfig}
\usepackage{colortbl}
\usepackage{booktabs}

\newcolumntype{C}{>{\centering\arraybackslash}p{1.2cm}}

\title{Why Not Transform Chat Large Language Models to Non-English?}

\author[1]{Xiang~Geng}
\author[2]{Ming~Zhu}
\author[1]{Jiahuan~Li}
\author[1]{Zhejian~Lai}
\author[1]{Wei~Zou}
\author[1]{Shuaijie~She}
\author[2]{Jiaxin~Guo}
\author[2]{Xiaofeng~Zhao}
\author[2]{Yinglu~Li}
\author[2]{Yuang~Li}
\author[2]{Chang~Su}
\author[2]{Yanqing~Zhao}
\author[2]{Xinglin~Lyu}
\author[2]{Min~Zhang}
\author[1]{Jiajun~Chen}
\author[2]{Hao~Yang}
\author[1,+]{Shujian~Huang}
\address[1]{National Key Laboratory for Novel Software Technology, Nanjing University, Nanjing 210023, China}
\address[2]{Translation Services Center, Huawei Inc., Beijing 100085, China}

\corremail{huangsj@nju.edu.cn}
\fcssetup{
  doi = {10.1007/s11704-025-50646-z},
  received = {month May, 2025},
  accepted = {month Aug, 2025},
}

\begin{abstract}
Chat large language models (LLMs), fine-tuned from pre-trained models and optimized for alignment with human preferences, 
excel in following diverse instructions while maintaining consistency with human values. 
In this paper, we propose the TransLLM framework for transforming chat LLMs from English to other languages using publicly available resources. 
TransLLM employs the translation chain-of-thought (TCoT) technique, which transfers chat ability through inference-time computation. 
Specifically, for a query in the target language, TCoT guides the LLM to first generate an English query and response as intermediate transfer steps before producing the final response in the target language. 
We underscore the necessity of improving the performance of each step in TCoT. % through continual pre-training (CPT).
However, improvement through continual pre-training (CPT) induces catastrophic forgetting of the original chat ability. %, even when low-rank adaptation (LoRA) is applied.  
To address this issue, we introduce recovery knowledge distillation (RKD), which utilizes data generated by the original chat LLM to recover its chat ability.
Experimental results indicate that TransLLM outperforms baseline models across various languages and LLMs while demonstrating adaptability in multilingual settings and generalizability beyond its training tasks.
Our analysis elucidates the mechanism by which RKD, in conjunction with LoRA, mitigates catastrophic forgetting. Code is available at \url{https://github.com/hy5468/TransLLM}.
\end{abstract}

\begin{document}

\section{Introduction}
\begin{figure*}[!ht]
\centering
\includegraphics[width=\textwidth]
{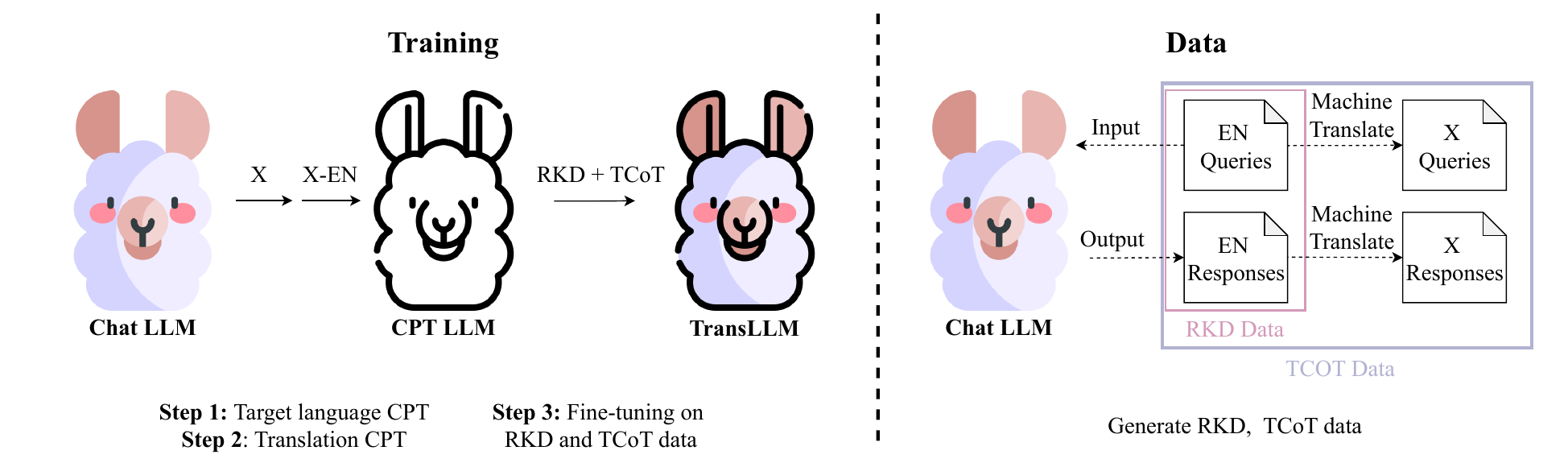}
\caption{
The pipeline of TransLLM.
% An illustration of how the chat LLM is transformed from English (EN) to the target language (X).
% The process begins by continual pre-training (CPT) the LLM on both X and EN monolingual corpora, as well as X-EN parallel data.
% Next, the LLM is fine-tuned using recovery knowledge distillation (RKD) and translation chain-of-thought (TCoT) data.
% Although the RKD and TCoT datasets are relatively small and focused on the instruction-following task, TransLLM can leverage them to transfer other capabilities, such as safety.
}
\label{fig:TransLLM}
\end{figure*}

Chat large language models (LLMs) such as GPT-4~\cite{gpt4} and Llama-3-Instruct~\cite{llama3} extend beyond their pre-trained counterparts (i.e., base LLMs) through supervised fine-tuning (SFT;~\cite{sft}) and preference optimization (PO;~\cite{RLHF,dpo}).
As a result, chat LLMs exhibit superior chat ability, enabling them to follow diverse instructions while aligning with human preferences.
However, since the majority of training data for chat LLMs is in English, their performance in non-English languages remains limited.
Even advanced chat LLMs, such as GPT-4, encounter safety challenges when tested with non-English languages~\cite{lrl_jailbreak}.

Moreover, the high-quality post-training English data is expensive and rarely made public, even when the corresponding chat LLMs are open-source.
Existing efforts~\cite{typhoon,aya_model,acegpt} to expand training data for target languages rely on open-source English data, but its quality and quantity significantly lag behind proprietary English datasets.
Since chat LLMs already possess extensive English knowledge from their proprietary datasets, why not transfer this knowledge to other target languages?

% Considering the huge effort in post-training, a more cost-effective way to obtain a chat LLM for a given language is to transform off-the-shelf chat LLMs from English to other languages, instead of training from scratch. 
% However, this transformation process faces several challenges.
% %Even with access to English chat data, the intricate training process remains impractical.
% First, 
% we need to transfer chat ability without STF and PO data in target languages.
% % while general chat ability encompasses a wide range of tasks, the available data may be limited to only a subset of tasks. Therefore, the proposed method should be generalizable to unseen tasks.
% %This discrepancy makes it challenging to transfer chat ability during training.
% Second, during continual learning, chat LLMs face a substantial risk of forgetting chat ability.
% Crucially, this issue cannot be mitigated by simply replaying~\cite{xuparadigm} SFT and PO procedures, as the required English SFT and PO data is both expensive and typically not made publicly available alongside model weights, as seen with Llama-3-Instruct~\cite{llama3}.

This paper addresses these issues by presenting a new framework named TransLLM, as illustrated in Figure~\ref{fig:TransLLM}.
TransLLM employs the strategy of translation chain-of-thought (TCoT;~\cite{plug}), which guides LLMs to generate a sequence of intermediate steps (i.e., chain-of-thought): 1. translating the query into English; 2. formulating a response in English; 3. generating the answer in the target language based on the preceding steps.
TCoT decomposes the transfer process into simple sub-tasks during inference, eliminating the need for SFT and PO data.
However, we underscore that the original TCoT presented in~\cite{plug} may not work due to the LLMs' insufficient performance on these sub-tasks. 

To address this question, TransLLM enhances performance in target language modeling (for step 3) and translation (for steps 1 and 3) via continual pre-training (CPT). 
Unfortunately, although CPT plays a critical role in language adaptation~\cite{chinese_Llama, typhoon}, it leads to catastrophic forgetting of the English chat ability, resulting in poor performance in step 2, even with the application of low-rank adaptation (LoRA; \cite{lora}).
% To mitigate catastrophic forgetting, we introduce recovery knowledge distillation (RKD), which utilizes data generated by the original chat LLM to recover its chat ability, in conjunction with LoRA.
To mitigate catastrophic forgetting, we propose recovery knowledge distillation (RKD), a technique that leverages data generated by the original chat LLM to recover its chat ability, in conjunction with LoRA.
% The insight is that fitting the RKD data with the original parameters is more straightforward than optimizing the LoRA parameters for the same task.
% This enables the LLM to learn a generalizable pattern that reduces the contribution of LoRA parameters when generating English content, thereby mitigating catastrophic forgetting.

% 强调human evaluation？
We conduct experiments across five languages (Thai, Arabic, Portuguese, Telugu, and Turkish) and three LLMs (Llama-2/3/3.1).
To thoroughly evaluate our method with limited resources, %we adopt a twofold strategy: an in-depth assessment of a single language and a broad evaluation across multiple languages. 
%Specifically, 
we first comprehensively evaluate TransLLM in Thai across various aspects, including instruction following, multi-turn conversation, commonsense reasoning, exam problem-solving, and safety. The evaluation is then extended for the instruction-following capability to the other four languages.
TransLLM significantly outperforms the reproduced baselines under comparable resources.
Compared to existing advanced baselines, TransLLM still demonstrates superior performance over both individual language LLMs (e.g., Jais~\cite{jais} and AceGPT~\cite{acegpt}) and multilingual LLMs (e.g., Llama-3.1~\cite{llama3} and Aya-101~\cite{aya_dataset}).
% Notably, our method rejects 94.61\% of harmful queries in Thai from the safety benchmark (AdvBenchmark).

Our main contributions are summarized as follows:
\begin{itemize}
    \item 
    We propose the TransLLM framework for transforming a chat LLM from English to other languages, enhancing both its helpfulness and safety in non-English contexts.
    \item 
   
    The experimental results indicate the effectiveness of TransLLM across various languages and LLMs.
    TransLLM can be easily adapted for multilingual use without compromising performance.
    % \item 
    % In TransLLM, we re-purpose existing techniques, TCoT and LoRA, to address the challenges of transforming chat LLMs.
    % Ablation studies show that existing techniques are ineffective without the TransLLM framework.
    \item 
    % While previous studies~\cite{} demonstrate that LoRA can mitigate catastrophic forgetting when training on small-scale datasets, we found it to be insufficient for large-scale training.
    The proposed RKD only requires fine-tuning on a small amount of data, without the need to cover all previously trained tasks.
    Our analysis demonstrates how RKD combined with LoRA helps mitigate catastrophic forgetting.
    \item 
    TransLLM allows non-English language performance to grow alongside the rapid development of English performance.
    We will make our code and datasets publicly available to facilitate further research in this area. 
\end{itemize}

% \begin{figure*}[t!]
% \centering
% \includegraphics[width=1\textwidth]{figures/pipeline.pdf}
% \caption{TransLLM pipeline.}
% \label{fig:pipeline}
% \end{figure*}

\section{Related Works}
\textbf{Language Adaptation in LLMs.} Previous studies have explored three primary directions:

(1) Adapting language during training.
Methods like Typhoon~\cite{typhoon}, Jais~\cite{jais}, and AceGPT~\cite{acegpt} first perform CPT on base LLMs using unlabeled data from target languages, followed by SFT with instruction-following data.
X-Llama~\cite{x_Llama} emphasizes semantic alignment through cross-lingual instruct tuning and translate training. 
Aya-101~\cite{aya_model} and Llama-3.1~\cite{llama3} contribute to advancing multilingual performance by leveraging a broad range of multilingual data.

(2) Adapting language during inference. 
Translate-bridge~\cite{translate-bridge, translate-bridge2} uses an additional translation model to bridge source and target languages, treating the TOCT steps as separate inference stages.
PLUG~\cite{plug} directly fine-tunes base LLMs using TCoT data. However, applying TCoT to chat LLMs presents challenges, as noted in previous discussions.

(3) Adapting language through model integration.
Recently, \cite{chat_vector} introduced the Chat Vector, which is derived by subtracting the weights of a base LLM from its chat version. 
This Chat Vector is then incorporated into the continuously pre-trained base LLM to endow it with chat ability.
However, since the Chat Vector is derived from English-focused LLMs, it may not generalize effectively to target languages, leading to suboptimal transfer performance.

\textbf{Mitigation of Catastrophic Forgetting in Continual Learning.} Previous methods can generally be classified into the following categories:

(1) Regularization-based methods~\cite{li2017learning,kim2023achieving} design regularization terms to preserve original knowledge during continual training.

(2) Parameter-based methods~\cite{serra2018overcoming,kang2022forget,ren2024analyzing} involve isolating parameters for different tasks during continual learning.
While LoRA has been shown to retain more of the original knowledge compared to full parameter training on small-scale datasets~\cite{ghoshcloser}, our findings suggest that it is insufficient for large-scale training.

(3) Replay-based methods~\cite{chaudhryefficient,sahagradient} replay previous training procedure on historical data. 
In parallel with our work, \cite{ssr} also use original LLMs to generate synthetic STF data for replay. However, we demonstrate that relying solely on RKD data is inadequate, as RKD cannot cover all previous knowledge.
As demonstrated in Section~\ref{sec:recover knowledge}, \emph{the key insight is that replaying with RKD data allows the model to isolate original parameters for original tasks by adaptively reducing the contribution of LoRA parameters.}

\section{Method}

\subsection{Model Architecture Extension}
TransLLM introduces two general extensions to the original LLM architecture, which are applicable to most existing LLMs.
\textbf{Vocabulary Expansion.} 
If the original tokenizer is trained on an English-dominated dataset, it causes over-fragmentation of non-English text. 
To address this, we extend the vocabulary for the target language to enhance model efficiency, as suggested by \cite{chinese_Llama,typhoon}.

\textbf{LoRA Modules.}
LoRA is a parameter-efficient training method commonly used in LLM training.
However, in this work, we utilize LoRA not only for efficiency but also to preserve the original model parameters.
Considering a weight matrix $W\in \mathbb{R}^{d\times k}$ of the original LLM and its input $h$, the output is $\Tilde{h} =Wh$.
LoRA represents the parameter update $\Delta W$ using two low rank matrices $B\in \mathbb{R}^{d\times r}$ and $A\in \mathbb{R}^{r\times k}$, where $r$ is the pre-determined rank. 
The updated output $\hat{h}$ is expressed as:
\begin{align}\label{LoRA}
 \hat{h} &= \Tilde{h}+\Delta Wh=\Tilde{h}+BAh.
\end{align}
\emph{Note: To maintain the original model parameters $W$, we do not merge the LoRA modules into the main backbone until all training stages is finished.}

\subsection{Training}
As shown on the left of Figure \ref{fig:TransLLM}, TransLLM employs a three-stage training process: target language CPT, translation CPT, and fine-tuning with RKD and TCoT data.
\subsubsection{Target Language Continual Pre-training}
Under-representation of the target language in the original LLMs leads to inadequate modeling of that language. 
Effective target language modeling is crucial for generating fluent, localized text and enhancing translation quality~\cite{mgnmt,llm_mt}.
To establish a strong foundation for the target language, we continually pre-train the LLM using monolingual data in the target language.
We do not introduce any English data at this stage due to the high cost of determining an optimal mixing ratio.

\subsubsection{Translation Continual Pre-Training}
TCoT leverages translation to bridge the gap between English and the target language.
Therefore, we introduce translation CPT to improve the bidirectional translation quality between English (EN) and the target language (X).
Inspired by mBART~\cite{mBART}, we use a special language ID token, such as $\langle\text{EN}\rangle$ and $\langle\text{X}\rangle$, to indicate the language transition within the text.
These tokens guide the behavior of TransLLM and facilitate its extension to multilingual settings, akin to mBART.
Given that sentences $s^{\text{EN}}$ from English and $s^{\text{X}}$ from language X are parallel, we treat this pair as two instances from different directions:
$(s^{\text{EN}},\langle\text{X}\rangle,s^{\text{X}})$ and $(s^{\text{X}},\langle\text{EN}\rangle,s^{\text{EN}})$, where $(\cdot,\cdot)$ denotes the concatenate operation. 

To restore fundamental English knowledge, we also incorporate English monolingual data for replay.
Specifically, we insert the translation instance between English monolingual data using $\langle\text{EOS}\rangle$ token as the separator.

\subsubsection{Transformation Fine-Tuning}
The CPT improves target language modeling and bidirectional translation performance. 
However, it also results in catastrophic forgetting of the original chat ability. 
During fine-tuning, our goal is to recover the LLM's chat ability while teaching the LLM how to execute TCoT.

\textbf{Recovery Knowledge Distillation.}
Previous studies have primarily focused on transforming base LLMs. To endow these base LLMs with instruction-following capabilities while using limited resources, a practical approach is to generate knowledge distillation (KD) data for instruction-following from advanced chat LLMs~\cite{alpaca}.
However, relying solely on limited KD data is insufficient for building strong chat ability.
To address this gap, we introduce the RKD method, which distills original knowledge from the chat LLM to help the CPT LLM recover its chat ability, rather than introducing new knowledge.
As shown on the right of Figure~\ref{fig:TransLLM}, for an English query $q^{\text{EN}}$, we use the original chat LLM to generate its corresponding answer $a^{\text{EN}}$.
We then introduce a special token $\langle \text{response} \rangle$ to guide the LLM that it should respond using its original chat ability.
As a result, the input and desired output for RKD fine-tuning are formulated as $q^{\text{EN}}$ and $(\langle \text{response} \rangle, a^{\text{EN}})$, respectively.

\textbf{Translation
Chain-of-Thought.}
TCoT simulates how humans handle second language questions by first interpreting and solving the question in their native language, then generating the response in the target language. 
As discussed in Section \ref{sec:better than translation}, the intermediate steps in TCoT contribute to the final answer, unlike in the translate-bridge approach, where each intermediate step is treated independently.
As shown on the right of Figure~\ref{fig:TransLLM}, we begin by machine translating the RKD data—$q^{\text{EN}}$ and $a^{\text{EN}}$—into the target language, resulting in $q^{\text{X}}$ and $a^{\text{X}}$, as done in previous works~\cite{x_Llama, typhoon}. 
Next, we organize the input and output of TCoT data as $q^{\text{X}}$ and $(\langle\text{EN}\rangle, q^{\text{EN}},\langle\text{response}\rangle, a^{\text{EN}},\langle\text{X}\rangle, a^{\text{X}})$.
In contrast to PLUG~\cite{plug}, we use special tokens rather than natural language to guide the model's behavior. This approach ensures that the special tokens do not interfere with the meaning of the natural language and facilitates easier extraction of results.

Due to the TCoT data, the model might struggle with accurately interpreting the translation instructions in X.
To address this, we also construct instruction-following data for translation by using translation prompt templates and parallel data. 
Finally, we combine all the aforementioned data randomly and fine-tune the CPT LLM.

\subsection{Inference}
The final TransLLM model can automatically respond 
EN instructions in EN, X instructions in TCoT mode, translation instructions in translation mode.
To fully leverage the original LLM’s multi-turn conversational ability in the target language, we strictly adhere to its native multi-turn format. 
For multi-turn tasks in the target language, only the English portions of the previous TCoT output are used as conversation history.
To be specific, we organize the input as $(q^{\text{EN}}_1,a^{\text{EN}}_1,\dots,q^{\text{EN}}_n,a^{\text{EN}}_n,q^{\text{X}}_{n+1})$, where $n$ is the number of past turns. 
We do not use any special tokens in the history as the original LLM does.
\emph{Interestingly, even in this unseen setting, the model still outputs the TCoT format as} $(\langle\text{EN}\rangle, q^{\text{EN}}_{n+1},\langle\text{response}\rangle, a^{\text{EN}}_{n+1},\langle\text{X}\rangle, a^{\text{X}}_{n+1})$.
The complete multi-turn template for Llama-2 is included as an example in Appendix Table~\ref{tab:inference template}.

\section{Experiments}

\subsection{Experiment Setup}\label{sec:settings}

\textbf{Settings.} 
Our experiments involve five languages, each with distinct resource availability and scripts: Thai (TH, mid-resource, Thai script), Arabic (AR, high-resource, Arabic script), Portuguese (PT, mid-resource, Latin script), Telugu (TE, low-resource, Telugu script), and Turkish (TR, mid-resource, Latin script).
We establish three experimental settings for different targets: 

(1) Transform Llama-2-Chat-7B~\cite{llama2} to TH (Llama-2 $\rightarrow$ TH). 
We reproduce strong baselines using comparable resources and the same backbone (i.e. Llama-2) to evaluate the effectiveness of TransLLM under fair conditions.
A comprehensive evaluation is conducted across multiple tasks, accompanied by a thorough analysis in this primary setting.

(2) Transform Llama-3-Instruct-8B~\cite{llama3} to AR (Llama-3 $\rightarrow$ AR).
This setting examines the generalizability of TransLLM for different languages and LLMs. 

(3) Transform Llama-3.1-Instruct-8B~\cite{llama3} to all five languages (Llama-3 $\rightarrow$ MTL).
This task evaluates the flexibility of TransLLM in the multilingual setting.

Due to limited resources, we only conducted settings (1) and (2) for TH and AR, respectively.

\textbf{Baselines.} 
The reproduced baselines encompass all three categories of adaptation methods, including training-time adaptation methods: X-Llama~\cite{x_Llama} and Typhoon~\cite{typhoon}; inference-time adaptation methods: PLUG~\cite{plug} and Translate-Bridge; model integration methods: Chat Vector~\cite{chat_vector}.
Detailed reproduction information can be found in Appendix A.

For the off-the-shelf models, we include Jais~\cite{jais} and AceGPT as individual language baselines, Aya-101~\cite{aya_model} and Llama-3.1-Instruct-8B~\cite{llama3} as multilingual baselines; ChatGPT~\cite{ChatGPT} and GPT-4~\cite{gpt4} as top-performing close-source baselines\footnote{The versions of ChatGPT and GPT-4 used are gpt-3.5-turbo-0125 and gpt-4-0613, respectively.}.

\textbf{Implementation.}
We implement our models using Chinese-Llama-Alpaca-2\footnote{\url{https://github.com/ymcui/Chinese-Llama-Alpaca-2}} and Llama-Factory \footnote{\url{https://github.com/hiyouga/Llama-Factory}} projects for Llama-2 and Llama-3/3.1,  respectively.
We only expand the vocabulary for Llama-2, as the vocabulary for Llama-3 is already optimized for multilingual use.
Special tokens, including language ID tokens and the $\langle \text{response} \rangle$ token, are added to the vocabulary.
We apply LoRA to the attention modules and multi-layer perceptron blocks, with the LoRA rank set to $r=64$. Further implementation details can be found in Appendix A.

\textbf{Training Data.}
For the CPT, we collect 11B and 30B tokens of TH and AR monolingual data from mC4~\cite{mc4}.
We sample 1M parallel pairs for EN-TH and EN-AR from CCAligned~\cite{CCAligned}, Tatoeba Challenge Data~\cite{tatoeba}, and OpenSubtitles~\cite{opensubtitles}; along with 1M EN documents from the Pile dataset~\cite{pile}.
To investigate whether TransLLM can operate effectively with constrained resources, no monolingual or parallel data is used for PT, TE, and TR.
For the fine-tuning, we use queries from the Alpaca dataset and generate corresponding responses using original chat LLMs for RKD.
We further use Google Translate to obtain TCoT data based on RKD data.
The Alpaca dataset contains a total of 52K queries, with the first 50K used as the training set and the remaining 2K reserved for validation in our experiments.
Detailed training procedures are provided in Appendix A.

\textbf{Benchmark.}
For TH, we use the following benchmarks:  multi-turn conversation and instruction-following benchmark MT-Bench~\cite{llm-as-a-judge}, instruction-following benchmark AlpacaEval~\cite{alpaca_eval}, causal commonsense reasoning benchmark XCOPA~\cite{xcopa}, exam problem-solving benchmark M3Exam~\cite{M3exam} and safety benchmark AdvBench~\cite{advbench}.
For AR, we use MT-Bench.
For PT, TE, and TR, we use the human-annotated Aya-Test~\cite{aya_dataset} for evaluation of instruction following.

We employ professional translators to translate MT-Bench and AlpacaEval from EN to target languages, similar to the creation of XCOPA.
Following the setting in~\cite{lrl_jailbreak}, we directly use Google Translate to translate the AdvBench.
Both M3Exam and Aya-Test are created by native speakers of the respective languages.

\begin{table*}[h!]
\centering
% % \vspace{-5pt}
\footnotesize
% \resizebox{\textwidth}{!}
{\begin{tabular}{c | c | c | c | c | c | c |  c | c}

\toprule
\multirow{2}{*}{Comparison} & \multicolumn{4}{c|}{First Turn (1st) }&  \multicolumn{4}{c}{Second Turn (2nd) } \\
& Win & Tie & Loss & $\Delta$ & Win & Tie & Loss & $\Delta$ \\
\midrule
\multicolumn{9}{c}{Results of Reproduced Models}\\
\midrule
% & PolyLM~\cite{Polylm} & 78.75 & 16.25 & \, 5.00 & \,\textbf{73.75} & 90.00 & \, 10.00 & \, 0.00 & \,\textbf{90.00} \\
TransLLM vs. X-Llama & 72.50 & 17.50 & 10.00 & \,\textbf{62.50} & 85.00 & \, 8.75 & \, 6.25 & \,\textbf{78.75}\\
TransLLM vs. Typhoon & 75.00 & 18.75 & \, 6.25 & \,\textbf{68.75} & 62.50 & 30.00 & \, 7.50 & \,\textbf{55.00}\\
TransLLM vs. PLUG & 72.50 & 13.75 & 13.75 & \,\textbf{58.75} & 87.50 & \, 8.75 & \, 3.75 & \,\textbf{83.75} \\
TransLLM vs. Translate-Bridge & 75.00 & 16.25 & \, 8.75 & \,\textbf{66.25} & 63.75 & 18.75 & 17.50 & \,\textbf{46.25} \\
TransLLM vs. Chat Vector& 78.75 & \, 8.75 & 12.50 & \,\textbf{66.25} & 85.00 & 13.75 & 1.25 & \,\textbf{83.75} \\
\midrule
\multicolumn{9}{c}{Results of Existing Models}\\
\midrule
TransLLM vs. ChatGPT~\cite{ChatGPT} & 42.50 & 26.26 & 31.25 & \,11.25 & 42.50 & 22.50 & 35.00 & \, 7.50  \\
TransLLM vs. GPT4~\cite{gpt4} & 26.25  & 28.75 & 45.00 & -18.75\, & 30.00 & 18.75 & 51.25 & -\textbf{21.75} \,\\
\bottomrule
\end{tabular}}
\caption{
Comparison of TransLLM (Llama-2 $\rightarrow$ TH) with various methods on  MT-Bench in TH. 
}
\label{tab:MT-Bench TH gpt-4 rate}
\end{table*}

\textbf{Evaluation.}
For the conversation benchmarks (MT-Bench, AlpacaEval, and Aya-Test), we primarily use GPT-4 for evaluation, following the procedure outlined in LLM-as-a-judge~\cite{llm-as-a-judge}.
Specifically, GPT-4 rates each response on a scale from 1 to 10. We then compute the win, tie, and loss rates by comparing the evaluation scores across different models.
The metric $\Delta$ represents the difference between the Win and Loss rates, calculated by including Tie annotations.
\emph{A larger $\Delta$ indicates that more of TransLLM's responses are preferred over those of the target competitor.}
To verify that the GPT-4 evaluation is reliable for non-English languages as well, we conduct human evaluations on several high-performing models and report the agreement between human and GPT-4 evaluations.

For choice-based benchmarks like XCOPA and M3Exam, we use Accuracy as the metric.

For the safety benchmark AdvBench, we track the rate of Bypass, Reject, and Unclear responses to harmful queries, as assessed by human evaluators.

In Appendix A, we describe the evaluation procedure, the instructions for human evaluators, and the information of evaluators in detail.
We also conduct significant tests for main results as described in Appendix~\ref{sec:statistical methods}.
We mark the results with \textbf{bold} if the difference is statistically significant ($p<0.05$).

\subsection{Main Results}

\textbf{Notable enhancement in multi-turn conversational and instruction-following ability. }% on target language
As shown in Table~\ref{tab:MT-Bench TH gpt-4 rate}, TransLLM significantly outperforms the reproduced baselines on MT-Bench, using comparable resources and the same backbone. 
A comparison between X-Llama and Typhoon reveals that the available CPT and SFT data offer limited knowledge in the target language. 
Similarly, the base LLM provides insufficient knowledge for PLUG to transfer, and transforming the chat LLM is not straightforward, as demonstrated in Section~\ref{sec:ablation studies} and~\ref{sec:recover knowledge}.
Although Translate-Bridge leverages additional translation resources, it remains inefficient, a point we discuss further in Section~\ref{sec:better than translation}. 
Additionally, the English-centered Chat Vector proves suboptimal for other languages, occasionally generating responses exclusively in English.

% Likewise, TransLLM surpasses strong baselines on both Alpaca-Eval 

\begin{table}[]
\centering
% % \vspace{-5pt}
\footnotesize
% \resizebox{\columnwidth}{!}
{\begin{tabular}{c |c | c | c | c }
\toprule
Lang & Model & Bypass  & Reject  & Unclear  \\
\midrule
\multirow{4}{*}{TH} & ChatGPT & 10.96 & 79.81 & 9.23 \\
&GPT4$^\dagger$  & 10.38 & 85.96 & 3.66 \\
&Ours w/ GPT-4 KD & 31.15 & 63.46 & 5.38 \\
&Ours & \, \textbf{2.69} & \textbf{94.61} & 2.69 \\
\midrule
\multirow{2}{*}{EN}&Llama-2 & \, 0.58 & 99.23 & 0.19 \\
&GPT4$^\dagger$ & \, 0.96 & 99.04 & 0.00 \\
\bottomrule
\end{tabular}}
\caption{
 Result on safety benchmark AdvBenchmark in the Llama-2 $\rightarrow$ TH setting under human evaluation. $^\dagger$ GPT-4 results are from \cite{lrl_jailbreak}.
}
% % \vspace{-15pt}
\label{tab:AdvBenchmark}
\end{table}

\begin{table*}[h!]
\centering
% % \vspace{-5pt}
\footnotesize
% \resizebox{\textwidth}{!}
{\begin{tabular}{c | C | C | C | C | C | C | C }

\toprule
\multirow{2}{*}{Comparison} & \multicolumn{2}{c|}{MT-Bench 1st $\Delta$ }& \multicolumn{2}{c|}{MT-Bench 2nd $\Delta$ }&  \multicolumn{3}{c}{Aya-Test $\Delta$ } \\
& TH & AR & TH & AR & PT & TE & TR \\
\midrule
TransLLM vs. Aya-101~\cite{aya_model} & 65.00 & 62.50 & 76.25 & 78.00 & \textbf{60.40} & \textbf{42.80} & \textbf{58.00} \\
TransLLM vs. Llama-3.1$\dagger$~\cite{llama3} & \textbf{31.25} & \textbf{32.50} & \textbf{43.75} & \textbf{35.00} & \textbf{29.60} & \textbf{36.40} & \textbf{49.60} \\
TransLLM vs. ChatGPT~\cite{ChatGPT} & \textbf{37.50} & \textbf{32.50} & \textbf{36.25} & \textbf{30.00} & \textbf{14.80} & \textbf{40.00} & \textbf{16.00} \\
TransLLM vs. GPT-4~\cite{gpt4} & \textbf{32.50} & \textbf{-31.25}\; & 17.50 & \textbf{-43.75}\; & \textbf{-29.20}\; & \textbf{-30.80}\; & \textbf{-34.00}\; \\
\bottomrule
\end{tabular}
}
\caption{
Comparison of TransLLM (Llama-3.1 $\rightarrow$ MTL) with various methods. $\dagger$ We additionally train Llama-3.1 using our training data (replacing TCOT data with translated RKD data) to rule out the possibility that our improvement is attributable to the use of extra data.
% To ensure a fair comparison in Table~\ref{tab:MT-Bench gpt-4 rate Llama-3.1}, 
}
% \vspace{-10pt}
\label{tab:MT-Bench gpt-4 rate Llama-3.1}
\end{table*}

\begin{table}[t]
\centering
% % \vspace{-10pt}
% \resizebox{\columnwidth}{!}
{
\begin{tabular}{c|c|c}
\toprule
          Comparison       & 1st $\Delta$  & 2nd $\Delta$  \\
\midrule
Llama-3.1 $\rightarrow$ MTL vs. Llama-3.1 $\rightarrow$ AR            &        0.00      &     -3.75          \\
Llama-3.1 $\rightarrow$ MTL vs. Llama-3 $\rightarrow$ AR         &  22.50& 25.00\\  
\bottomrule
\end{tabular}}
\caption{
Comparison of TransLLM (Llama-3.1 $\rightarrow$ MTL) with other TransLLM configurations on MT-Bench in AR.
}
% \vspace{-15pt}
\label{tab:TransLLM variants}
\end{table}

% \textbf{Higher safety than widely used ChatGPT and GPT-4.} 
\textbf{Making LLMs safe in non-English languages.} 
Due to the cost of human evaluation, we only compare TransLLM with widely used ChatGPT and GPT-4, which have been specifically optimized for safety.
In Table \ref{tab:AdvBenchmark}, TransLLM has a rejection rate of 94.61\%, close to 99.23\% of the original LLM.
It indicates that we successfully transfer most of the human preference about the safety of the original LLM.
TransLLM attains an improvement of 14.8\% and 8.65\% over ChatGPT and GPT-4 for rejecting harmful queries with statistical significance.
More importantly, while GPT-4 is safe in EN, replacing RKD data (including the RKD portion of the TCoT data) with GPT-4 KD data may increase harmful responses.
Later, we will demonstrate that this is because RKD recovers the original knowledge.

\textbf{Generalizability across various settings.}
As shown in Appendix B,
TransLLM also achieves superior performance on several other benchmarks, including Alpaca-Eval (Table~\ref{tab:Alpaca-Eval gpt-4 rate}), XCOPA (Table~\ref{tab:accuracy XCOPA}), and M3Exam (Table~\ref{tab:accuracy M3Exam}) in TH.
In the Llama-3 $\rightarrow$ AR setting (Table~\ref{tab:MT-Bench AR gpt-4 rate}), TransLLM still demonstrate competitive performance compared to Jais and AceGPT, both of which are advanced LLMs specifically developed for individual language AR.
Notably, TransLLM outperforms ChatGPT with publicly available resources using fewer parameters in all settings. 

% \begin{table*}[h!]
% \centering
% % \vspace{-5pt}
% \footnotesize
% \resizebox{\textwidth}{!}
% {\begin{tabular}{c |c |c | c | c | c | c }

% \toprule
% \multirow{1}{*}{Llama-3.1 to}& Benchmark & \multirow{1}{*}{TransLLM vs.}  & Win & Tie & Loss & $\Delta$  \\
% \midrule
%  & \ &  &  &  &  & \\
% \bottomrule
% \end{tabular}}
% \caption{
% Comparison between our model and different methods on MT-Bench in TH and AR under GPT-4 evaluation.
% }
% % \vspace{-15pt}
% \label{tab:MT-Bench gpt-4 rate}
% \end{table*}
\subsection{Multilingual Results}

\textbf{Surpass strong multilingual baselines.} 
Table \ref{tab:MT-Bench gpt-4 rate Llama-3.1} shows that TransLLM consistently outperforms Llama-3.1$\dagger$—which utilizes the same resources and backbone—across all five languages.
Additionally, TransLLM demonstrates superior performance compared to the open-source multilingual LLM Aya-101 and the commercial LLM ChatGPT.

\textbf{Consistent improvement in target languages, regardless of initial proficiency.} 
On one hand, results in Table~\ref{tab:MT-Bench TH gpt-4 rate} demonstrate that TransLLM successfully develops TH capabilities for Llama-2, despite Llama-2 lacking native support for the TH language.
On the other hand, Table \ref{tab:MT-Bench gpt-4 rate Llama-3.1} demonstrates that TransLLM continues to enhance the TH  performance of Llama-3.1, even though Llama-3.1 already possesses some proficiency in TH.

\textbf{Multilingual TransLLM achieves similar performance to vanilla TransLLM.} We also transform Llama-3.1 to the individual language AR. 
As shown in Table~\ref{tab:TransLLM variants}, the performance of multilingual and vanilla TransLLM models is comparable.
This phenomenon can be attributed to the effectiveness of language IDs in helping models differentiate between tasks in various languages.

\textbf{Stronger chat LLMs lead to better performance.} 
From Table \ref{tab:TransLLM variants}, we observe that the performance of TransLLM consistently improves when stronger chat LLMs are used.
The continuous development of increasingly powerful LLMs highlights the growing potential of TransLLM.

\begin{table*}[h]
\centering
% % \vspace{15pt}
\footnotesize
{\begin{tabular}{c | c | c | c | c | c | c | c |  c | c}

\toprule
\multirow{2}{*}{Setting} & \multirow{2}{*}{Comparison} & \multicolumn{4}{c|}{First Turn } & \multicolumn{4}{c}{Second Turn } \\
& & Win & Tie & Loss & $\Delta$ & Win & Tie & Loss & $\Delta$ \\
\midrule
\multirow{2}{*}{Llama-2 $\rightarrow$ TH}& TransLLM vs. ChatGPT & 53.75 & 27.50 & 18.75 & \, \textbf{35.00} & 48.75 & 26.25 & 25.00 & \, \textbf{23.75} \\
& TransLLM vs. GPT-4 & 22.50 & 40.00 & 37.50 & -15.00 & 22.50 & 27.50 & 50.00 & \,\,\textbf{-27.50}\;\\
\midrule
\multirow{2}{*}{Llama-3 $\rightarrow$ AR}& TransLLM vs. ChatGPT & 50.00 & 30.00 & 20.00 & \,\textbf{30.00} & 42.50 & 35.00 & 22.50 & \,\,\textbf{20.00} \\
& TransLLM vs. GPT-4 & 17.50  & 30.00 & 52.50 & \textbf{-35.00}\, & 12.50  & 27.50 & 60.00 & \, \textbf{-47.50} \,\\
\bottomrule
\end{tabular}}
\caption{
Comparison of TransLLM with advanced LLMs on MT-Bench under human evaluation.
}
% % \vspace{-15pt}
\label{tab:MT-Bench huamn eval rate}
\end{table*}

% \subsection{Agreement Results}
\textbf{High agreement between humans and GPT-4 evaluation.}
We perform human evaluation on MT-Bench in TH and AR for TransLLM, ChatGPT and GPT-4 as shown in Table~\ref{tab:MT-Bench huamn eval rate}. 
Then, we compute the average agreement rates by comparing each pair of models.
In Appendix Table~\ref{tab:human evaluation agreement},  GPT-4 shows high consistency with human annotators.
The consistency between GPT-4 and humans is much higher than random guesses and comparable with the consistency in EN.
\cite{aya_model} show high agreement between humans and GPT-4 evaluation on Aya-Test.
Therefore, we use GPT-4 to evaluate MT-bench and Aya-Test in this paper. 
% Due to resource limitations, we do not perform human evaluations for PT, TE, and TR.

\section{Analysis}

\begin{table}[t]
\centering
% % \vspace{-10pt}
% \resizebox{\columnwidth}{!}
{
\begin{tabular}{c|c|c|c}
\toprule
 & TransLLM vs.                 & 1st $\Delta$  & 2nd $\Delta$  \\
\midrule
1 &  -Chat LLM  +Base LLM          & 36.25             & 67.50              \\
2 & -TCoT -RKD +TH SFT            & 46.25                 & 47.50              \\ 
3 & -TCoT +TH SFT            & 41.25                 & 45.00              \\ 
\midrule
4 & -TH CPT          & 41.25             & 35.00              \\
5 & -Translation CPT & \, 8.75             & 23.75              \\
6 & -RKD +GPT-4 KD               & 17.50             & 45.00              \\
7 & -LoRA in CPT                 & 41.25             & 46.25
              \\
8 & -LoRA in CPT and SFT                & 62.50             & 66.25              \\
\midrule
9 &-Special Tokens +NLT Language & 23.75             & 7.50              \\
10 &-TCoT w/ Google +TCoT w/ NLLB  & 17.50             & 35.00              \\
11 & -EN history +TH history             & -                 & 23.75              \\ 
\bottomrule
\end{tabular}}
\caption{
Comparison of TransLLM (Llama-2 $\rightarrow$ TH) with ablation models on MT-Bench. ``-'' and ``+'' indicates the removal and addition of the element, respectively.
}
% \vspace{-15pt}
\label{tab:ablation studies delta}
\end{table}

\subsection{Ablation Studies}\label{sec:ablation studies}
We conduct comprehensive ablation studies to investigate the impact of TransLLM's components.
Based on results in Table~\ref{tab:ablation studies delta}, we have the following observations.

\textbf{Transforming chat LLMs through TCoT offers significant improvements.} 
Chat LLMs offer better chat ability compared to base LLMs (Line 1). However, directly SFT with instruction-following data in TH does not fully transfer chat ability (Line 2), even when chat ability is recovered (Line 3).

\textbf{The performance of intermediate steps is crucial for the success of TCoT.}
% TransLLM without TH CPT is less satisfying.
Without CPT on TH documents, the LLM is less satisfying for TH language modeling (Line 4).
Since TH CPT and transformation fine-tuning also improve translation ability, the improvement of the translation CPT is not as significant as other components (Line 5).
Beyond safety, the high-quality GPT-4 KD data also leads to performance degradation in conversation (Line 6).
That is because our goal is not to inject more knowledge but to recover the original knowledge for TCoT.
We also analyze the contribution of LoRA, as detailed in Appendix B. 
In most tasks, full fine-tuning performs similarly to or better than LoRA. 
However, in our case, full fine-tuning erases the original knowledge encoded in the parameters, leading to a significant drop in performance compared to TransLLM with LoRA, both during the CPT (Line 7) and the entire training procedure (Line 8).

\textbf{The impact of template design, data quality, and chat history on TCoT.}
% TransLLM without TH CPT is less satisfying on both the first and second turn (Line 2).
% Pre-training on TH documents helps TransLLM output fluency in TH response with long context.
% Thus, TransLLM without TH pre-training is less satisfying on both the first and second turn (Line 2).
% Since TH pre-training and transformation fine-tuning also provide some translation knowledge, the improvement of the translation pre-training is not as significant as other components (Line 3).
% Beyond safety, the high-quality GPT-4 KD data also leads to performance degradation in helpfulness (Line 4).
% That is because our goal is not to inject more knowledge but to preserve the original knowledge.
% We also examine the contribution of LoRA.
% Specifically, we merge the LoRA parameters with full parameters before transformation fine-tuning.
% We are unable to conduct full fine-tuning for per-training, but the merged model is a good approximation according to Eq.~\ref{LoRA}.
% We further conduct transformation fine-tuning with full parameters based on the merged model.
% In most tasks, full fine-tuning is better or comparable with LoRA.
% However, in our case, full fine-tuning wipes the original knowledge from parameters, and therefore its performance is much lower than TransLLM with LoRA (Line 5).
The natural language template (NLT) \footnote{``Let me interpret the instruction in English:...Then the English response is:...Finally, the Thai response is:...''} for TCoT, as used in \cite{plug}, slightly reduces performance (Line 9).
Repeated training may have altered the original meaning of natural language in the template.
The TCoT data generated by Google Translate achieves better performance than that produced by NLLB (Line 10), as Google Translate provides higher translation quality, reflected in its CometKiwi~\cite{cometkiwi} score of 83.40 compared to NLLB's 79.07.
When using the history in TH, TransLLM is also capable of multi-turn conversation with small performance degradation (Line 11). 
% That means TransLLM can handle TH context well, this ability could be further developed for retrieval augmentation in TH.

% \subsection{TransLLM Recover the Original Knowledge}\label{sec:recover knowledge}
\subsection{Catastrophic Forgetting}\label{sec:recover knowledge}

\textbf{Knowledge is forgotten and recovered.} 
To measure how much original knowledge is forgotten by the chat LLM, we calculate the average difference between the generation probabilities of the original LLM and different models, using the hold-out validation set of RKD data.
As shown in Table~\ref{tab:generation probabilities}, the LLM significantly forgets the original knowledge after CPT (Line 2).
While the Chat Vector preserves some knowledge, it tends to generate more English responses when queried in TH (Line 3).
% Following the recommendation of \cite{chat_vector}, we addressed this issue by reducing the Chat Vector weight to 0.5, resulting in less knowledge retention (Line 4).
RKD (Line 4) and LoRA (Line 5) are both essential for knowledge retention.
TransLLM, with a probability of 0.2352 and a difference of 0.0055, demonstrates almost no loss of knowledge compared to other models, indicating it retains nearly all the original knowledge (Line 6).

\textbf{Mechanism of mitigating catastrophic forgetting.}
During CPT, new knowledge is incorporated into the LoRA parameters.
As demonstrated in Eq. \ref{LoRA}, although the original parameters remain unchanged, their representation is altered by LoRA parameters, leading to a significant degradation in chat ability.
When fine-tuning on RKD data,
LLM can fit the RKD data easily by reducing the contribution of LoRA parameters.
% 同时我们需要LoRA参数在处理TH时利用新学到的知识发挥作用
We assume that may enable the LLM to learn a generalizable pattern that uses the original knowledge for EN and new knowledge for others.

To test this assumption, we calculated the cosine similarity between the last layer's hidden states of the original model ($\Tilde{h}$ in Eq.~\ref{LoRA}) and the LoRA-updated model ($\hat{h}$ in Eq.~\ref{LoRA}), using TCoT validation data.
Higher similarity indicates greater reliance on original knowledge.
The average similarity per token for English responses (0.6191) is significantly higher than for TH responses (0.2522).
The result indicates that TransLLM effectively learns this pattern by using LoRA and RKD together.
% The insight is that using the original parameters to fit the RKD data is much easier than optimizing the LoRA parameters from scratch.
% % This enables the LLM to learn a pattern that improve the contribution of original parameters when generating English content, thereby mitigating catastrophic forgetting.
% This enables the LLM to learn a pattern that reduce the contribution of LoRA parameters when generating English content, thereby mitigating catastrophic forgetting.
% During the training process, 
% new knowledge is learnt into the LoRA parameters.

\subsection{TransLLM vs. Translate-Bridge}\label{sec:better than translation}

% \begin{table}{l}{\columnwidth}
% \centering
% \footnotesize
% % \vspace{-5pt}
% \resizebox{\columnwidth}{!}
% {\begin{tabular}{c | c | c | c | c }
% \toprule
% \multirow{2}{*}{Model} & \multicolumn{2}{c|}{EN-TH}  & \multicolumn{2}{c}{TH-EN}  \\
%  & COMET  &  BLEU & COMET  &  BLEU \\
% \midrule
% ChatGPT & 85.47 & 31.26 & 86.29 & 23.47 \\
% NLLB & 83.88 & 28.53 & 87.14 & 30.78 \\
% Ours & 86.96 & 35.04 & 86.97 & 27.68 \\
% \bottomrule
% \end{tabular}}
% \caption{
% Translation performance on Flores-200.
% }
% % \vspace{-10pt}
% \label{tab:translation performance}
% \end{table}

% \begin{table}{r}{0.35\textwidth}
% \centering
% \footnotesize
% % \vspace{-5pt}
% {\begin{tabular}{c | c }
% \toprule
% Model &   Score  \\
% \midrule
% Translate-Bridge & 5 \\
% GPT4 &6 \\
% ChatGPT & 7 \\
% Ours & 7 \\ 
% \bottomrule
% \end{tabular}}
% \caption{
% Fluency on MT-Bench.
% }
% % \vspace{-10pt}
% \label{tab:fluency}
% \end{table}
\begin{table}[]
\centering
\footnotesize
% % \vspace{-10pt}
% \resizebox{\columnwidth}{!}
{\begin{tabular}{c |c | c | c }
\toprule
&Model & $P(y|x)$ & Difference\\
\midrule
1&Llama-2 & 0.2363 & - \\
2& CPT LLM  & 0.1666 & 0.0697 \\
% 3&Chat Vector & 0.2118 & 0.0245 \\
3&Chat Vector & 0.1873 & 0.0490 \\
4&Ours w/ GPT-4 KD & 0.1972 &  0.0391 \\
5&Ours w/o LoRA & 0.1772 &  0.0592 \\
6&Ours  & 0.2352 & 0.0055 \\
\bottomrule
\end{tabular}}
\caption{
The difference of generation probabilities in the Llama2 $\rightarrow$ TH setting.
}
% \vspace{-15pt}
\label{tab:generation probabilities}
\end{table}

\textbf{Competitive translation performance.} The translation performance is critical for both TransLLM and Translate-Bridge.
Therefore, we test them on the widely used benchmark Flores-200~\cite{Flores-101}.
Benefiting from TH and translation CPT, TransLLM outperforms ChatGPT and NLLB~\cite{nllb} on EN-TH and achieves competitive performance on TH-EN (Appendix Table~\ref{tab:translation performance}).
We also ask the naive TH speaker to provide a fluency score for each model on MT-Bench.
Appendix Table~\ref{tab:fluency} demonstrate that TransLLM is generally fluent.
% The fluency of NLLB is as poor as its translation performance on EN-TH.
% NLLB usually translates the responses literally.
% For example, NLLB translates ``I see'' into ``I see something'' instead of ``I understand'' in TH. 
% Surprisingly, the response of GPT-4 is not very fluent and natural.
% GPT-4 often uses full-stops and commas which are not used in TH.
% ChatGPT and TransLLM are generally fluent, with translationese to a certain degree.
% For example, TH speakers do not use ``sure'' or ``of course'' at the beginning of responses, but ChatGPT and TransLLM do.

\textbf{TransLLM is more than translation.} %Translation performance is important but not the whole story.
%TransLLM outputs an EN query, EN response, and TH response at once.
% TransLLM can use all previous information for TH responses and therefore achieve better performance than Translate-Bridge.
% Since NLLB is not, 
We further use TransLLM as the translation model for Translate-Bridge.
The performance is still worse than TransLLM with $\Delta=$ 13.75\% and $\Delta=$ 18.75\% on first and second turn.
This indicates that TCoT is more effective than the Translate-Bridge approach, even when both possess the same translation performance.
The attention map of TransLLM in Appendix Figure~\ref{fig:attention} shows that TransLLM outputs the TH response mostly based on the TH response itself and then the EN response.
However, the TH response also pays a little attention to the TH query and EN query.
% Besides, Translate-Bridge needs to deploy two models, which is costly and inconvenient.

% Some concurrent works attempt to mi

\section{Conclusion}\label{sec: Conclusion}
% Chat LLMs have been specifically optimized for chat usage and therefore are helpful and safe in the dominant language.
In this paper, we propose a framework for transforming an off-the-shelf chat LLM to other languages.
In this framework, we utilize TCoT to transfer English capabilities and further enhance the performance of TCOT's intermediate steps.
To recover the original knowledge, we propose the RKD method supplemented with LoRA.
The experiments across different languages and LLMs show that TransLLM transfers desired capabilities to the target language and outperforms strong baselines in both helpfulness and safety.
% Overall, we hope this work can become the foundation for developing safe LLMs in many languages other than English.

\section*{Limitations}
Our experiments have some limitations due to resource constraints:
(1) We focused on LLMs with fewer than 8B parameters. The effects of scaling up parameters require further exploration.
(2) Our experiments were limited to the Llama series of LLMs. While Llama-2, Llama-3, and Llama-3.1 differ in training methodology, data, and architecture, testing with other LLM families would help assess the broader applicability of TransLLM.
(3) Human evaluation is currently limited to several models in TH and AR. In future work, we plan to expand the evaluation to include additional baselines and languages, ensuring a more comprehensive and reliable assessment.
% (3) Due to huge cost of CPT, we only reproduced baselines in the Llama-2 $\rightarrow$ TH setting. 

For now, TransLLM is still highly dependent on translation.
Consequently, TransLLM can not handle the queries related to language features.
A straightforward approach is to train TransLLM to determine whether to respond using TCoT mode or not.
Additionally, the translation resources may be lacking for extremely low-resource languages. We could explore unsupervised or semi-supervised machine translation techniques, such as iterative back-translation and multilingual pre-training, to improve translation performance without relying on parallel data.

Due to the TCoT, the inference overhead of TransLLM is much longer than other baselines.
Recently, \cite{Pause} and \cite{ICoT} show that the implicit chain-of-thought achieves similar performance on reasoning tasks without additional inference overhead.
We would like to explore TransLLM with implicit TCoT in the future.

\section*{Acknowledgement}
We would like to thank the anonymous reviewers for their insightful comments. Shujian Huang is the corresponding author. This work is supported by National Science Foundation of China (No. 62376116, 62176120), the Fundamental Research Funds for the Central Universities (No. 2024300507).

\section*{Competing Interest}
The authors declare that they have no competing interests or financial conflicts to disclose.

\begin{table}[h!]
\centering
\footnotesize
{\begin{tabular}{l}
\toprule
$\langle s \rangle[\text{INST}] \langle \langle \text{SYS} \rangle\rangle$\\
You are a helpful assistant.
$\langle\langle /\text{SYS} \rangle\rangle$\\
\\
$q^\text{EN}_1$ [/INST] $a^\text{EN}_1$ $\langle /s \rangle$\\
$\langle s \rangle$[INST] $q^\text{EN}_2$ [/INST] $a^\text{EN}_2$ $\langle /s \rangle$\\
...\\
$\langle s \rangle$[INST] $q^\text{EN}_n$ [/INST] $a^\text{EN}_n$ $\langle /s \rangle$\\
$\langle s \rangle$[INST] $q^\text{X}_{n+1}$ [/INST] \\
\bottomrule
\end{tabular}}
\caption{
The multi-turn prompt template used in our experiments.
}
\label{tab:inference template}
\end{table}

\begin{figure}[ht!]
\centering
\includegraphics[width=1.8in]{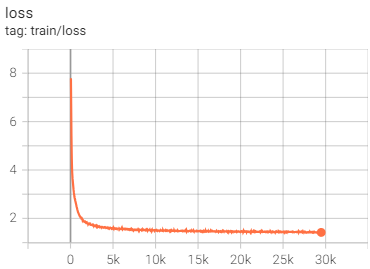}
\caption{Pre-Training loss for TH.}
\label{fig:TH pretrain loss}
\end{figure}
% \section*{Appendixes}

\begin{table*}[h!]
\centering
\footnotesize
% \resizebox{\textwidth}{!}
{\begin{tabular}{c | c | c | c | c }
\toprule
Name & LLM & Pre-train Data & Fine-tune Data & Size \\
\midrule
% PolyLM & From Scratch & MTL + Translation & Alpaca-MTL & 13B \\
X-Llama & Llama-2-base & - & Alpaca-EN + Alpaca-TH + Translation & 7B \\
Typhoon & Llama-2-base & TH & Alpaca-TH & 7B \\
PLUG & Llama-2-base & - & TCoT & 7B  \\
NLLB bridge & Llama-2-chat + NLLB &  - & - & 7B + 3B \\
Chat Vector & Llama-2-base/chat &  TH & Alpaca-TH & 7B \\
% ChatGPT & Unknown & Unknown & Unknown & $\gg$ 7B \\
% GPT4 & Unknown & Unknown & Unknown & $\gg$ 7B \\
Ours & Llama-2-chat & TH / Translation + EN & TCoT + RKD + Translation & 7B \\
\bottomrule
\end{tabular}}
\caption{
Details of the reproduced models in the Llama-2 $\rightarrow$ TH setting.
}
\label{Model details}
\end{table*}

\begin{table*}[h]
\centering
\footnotesize
{\begin{tabular}{c | c | p{0.65\textwidth} }
\toprule
Score & Performance Level & Adherence to Instructions; Expression Fluency; Style\\

\midrule
1-2   & Very Poor  & Does not follow the query; be not applicable due to 
nonsensical expression;
has incomprehensible style                                                                   \\
3-4   & Poor          & Does not follow the query but has some relevant content; lacks fluency, coherency, and clarity; has largely inappropriate style                      \\
5-6   & Fair & Partially meets the requirements and addresses some issues; has some fluency and clarity though minor flaws; has occasionally appropriate style\\
7-8   & Good          & Mainly follows the query though some minor flaws; be largely fluent and coherent; has generally appropriate style                                 \\
9-10  & Excellent       & Strictly follows the query with appreciated content; has a high degree of fluency and clarity; is perfectly matched in style                                                                       \\
\bottomrule
\end{tabular}}
\caption{
Rating criterion for human evaluation.
}
\label{tab:Annotation instruction}
\end{table*}

\begin{table*}[h!]
\centering
% \vspace{-5pt}
\footnotesize
{\begin{tabular}{c | c | c | c | c}
\toprule
\multirow{2}{*}{Setting} & \multicolumn{2}{c}{First Turn } & \multicolumn{2}{c}{Second Turn } \\
 & w/ Tie (R = 33\%) & w/o Tie(R = 50\%) & w/ Tie (R = 33\%) & w/o Tie (R = 50\%) \\
\midrule
Result & 75.00 & 91.70 & 67.50 & 80.00 \\
\bottomrule
\end{tabular}}
\caption{
Inter-annotator agreement on MT-Bench in TH.
}
% % \vspace{-15pt}
\label{tab:Inter-annotator agreement}
\end{table*}

\begin{table}[h!]
\centering
\footnotesize
\begin{tabular}{c|c}
\toprule
Resource & License \\
\midrule
MC4~\cite{mc4}                      & ODC-BY 1.0 \\
Pile~\cite{pile}                   & MIT License                           \\
CCAligned~\cite{CCAligned} & Unknown\\
Tatoeba Challenge Data~\cite{tatoeba} & CC-BY-NC-SA 4.0 \\
OpenSubtitles~\cite{opensubtitles} & Unknown \\
Flores-200~\cite{Flores-101} & CC-BY-SA 4.0 \\
Alpaca~\cite{alpaca}                    & CC BY-NC 4.0 \\
Alpaca-eval~\cite{alpaca_eval}            & Apache License 2.0                   \\
MT-Bench~\cite{llm-as-a-judge} & Apache License 2.0 \\
AdvBench~\cite{advbench} & MIT License \\
Aya-Test~\cite{aya_dataset} & Apache License 2.0 \\
M3Exam~\cite{M3exam} & Unknown \\ 
Chinese-Alpaca-2~\cite{chinese_Llama}           & Apache License 2.0 \\
Transformers~\cite{transformers}               & Apache License 2.0                   \\
SentencePiece~\cite{sentencepiece}              & Apache License 2.0                   \\
PolyLM~\cite{Polylm}                    & Apache License 2.0                   \\
Llama-2~\cite{llama2}                     & Llama 2 Community License Agreement \\
Llama-3~\cite{llama3}                     & Llama 3 Community License Agreement \\
\bottomrule
\end{tabular}
\caption{Licenses of open source resources.}
\label{License}
\end{table}

\begin{table}[h!]
\centering
% % \vspace{-10pt}
\footnotesize
% \resizebox{\columnwidth}{!}
{\begin{tabular}{c | c | c | c | c}

\toprule
TransLLM vs. & Win  &  Tie  & Loss  & $\Delta$  \\
\midrule
X-Llama & 92.50 & \, 5.00 & \, 2.50 & \textbf{90.00} \\
PLUG & 87.50 & \, 8.75 & \, 3.75 & \textbf{83.75} \\
Translate-Bridge & 91.25 & \, 5.00 & \, 3.75 & \textbf{87.50} \\
ChatGPT & 72.50 & 13.75 & 13.75 & \textbf{58.75} \\
GPT4 & 17.50 & 45.00 & 37.50 & \textbf{-20.00}\; \\
\bottomrule
\end{tabular}}
\caption{
Comparison between TransLLM (Llama-2 $\rightarrow$ TH) with different methods on Alpaca-Eval in TH. 
}
% \vspace{-5pt}
\label{tab:Alpaca-Eval gpt-4 rate}
\end{table}

\begin{table}[h]
\centering
\footnotesize
{\begin{tabular}{c | c}

\toprule
Model     & Acc \\
\midrule
Typhoon   & 36.20 \\
X-Llama   & 46.00 \\
Translate-Bridge & 55.20 \\
ChatGPT   & 51.70 \\
Ours  & 60.08 \\
GPT-4     & 75.20 \\
\bottomrule
\end{tabular}}
\caption{
Accuracy (\%) of different models  on XCOPA in the Llama-2 $\rightarrow$ TH setting.
}
\label{tab:accuracy XCOPA}
\end{table}

\begin{table}[h]
\centering
\footnotesize
{\begin{tabular}{c | c}

\toprule
Model     & Acc \\
\midrule
Random & 25.00  \\
Llama-3.1  & 28.09 \\
ChatGPT$^\dagger$   & 46.00 \\
TransLLM (Llama-3.1 $\rightarrow$ MTL)  & 46.68 \\
GPT-4$^\dagger$     & 56.04 \\
\bottomrule
\end{tabular}}
\caption{
Accuracy (\%) of different models  on M3Exam in TH. $^\dagger$ results are from \cite{llm-as-a-judge}.
}
\label{tab:accuracy M3Exam}
\end{table}

\begin{table*}[h]
\centering
% % \vspace{-5pt}
\footnotesize
% \resizebox{\textwidth}{!}
{\begin{tabular}{c | c | c | c | c | c | c |  c | c}

\toprule
\multirow{2}{*}{TransLLM vs.} & \multicolumn{4}{c|}{First Turn }&  \multicolumn{4}{c}{Second Turn } \\
& Win & Tie & Loss & $\Delta$ & Win & Tie & Loss & $\Delta$ \\
\midrule
Jais~\cite{jais} & 56.25 & 25.00 & 18.75 & \textbf{37.50} & 48.75 & 33.75 & 17.50 & \textbf{31.25} \\
AceGPT~\cite{acegpt} & 38.75 & 33.75 & 27.50 & 11.25 & 61.25 & 17.50 & 21.25 & \textbf{40.00} \\
ChatGPT~\cite{ChatGPT} & 45.00 & 22.50 & 32.50 & 12.50 & 46.25 & 26.25 & 27.50 & {18.75} \\
GPT-4~\cite{gpt4} & 12.50 & 37.50 & 50.00 & \textbf{-37.50} & \, 8.75 & 28.75 & 62.50 & \textbf{-53.75} \\
\bottomrule
\end{tabular}}
\caption{
Comparison of TransLLM (Llama-3 $\rightarrow$ AR) with various methods on  MT-Bench in AR.
}
% \vspace{-5pt}
\label{tab:MT-Bench AR gpt-4 rate}
\end{table*}

\begin{table*}[h!]
\centering
\footnotesize
% \resizebox{\textwidth}{!}
{\begin{tabular}{c | c | c | c | c| c | c}
\toprule
\multirow{2}{*}{Setting} & \multicolumn{2}{c|}{TH} & \multicolumn{2}{c|}{AR} &\multicolumn{2}{c}{EN$^\dagger$} \\
& First Turn  &  Second Turn  & First Turn  &  Second Turn  & First Turn  &  Second Turn   \\
\midrule
w/ Tie (R = 33) & 75.42 & 70.42 & 58.00 & 58.44 & 60.00 & 59.00 \\
w/o Tie (R = 50) & 75.11 & 67.85 & 87.36 & 87.04 & 85.00 & 84.00 \\
\bottomrule
\end{tabular}}
\caption{
Agreement between GPT-4 and humans. ``R='' denotes the expect agreement between random judges. $^\dagger$ EN results are from \cite{llm-as-a-judge}.
}
% \vspace{-5pt}
\label{tab:human evaluation agreement}
\end{table*}

\begin{table*}[]
\centering
\footnotesize
\resizebox{\textwidth}{!}{\begin{tabular}{c c | c | c | c | c | c | c | c | c | c}
\toprule
& Model & Writing & Roleplay & Reasoning & Math & Coding & Extraction & STEM & Humanities & All\\
\midrule
\multirow{3}{*}{{\shortstack{\textbf{First}\\ \textbf{Turn}}}} & ChatGPT & 5.30 & 4.70 & 5.20 & 4.60 & 7.80 & 7.20 & 6.80 & 6.40 & 6.00 \\
& GPT4 & 7.40 & 6.70 & 4.80 & 6.00 & 8.80 & 8.30 & 7.40 & 7.70 & 7.14 \\
& Ours (Llama-2 $\rightarrow$ TH) & 7.30 & 6.50 & 5.20 & 4.20 & 6.50 & 5.70 & 7.60 & 7.90 & 6.36 \\%6.3625
\midrule
\multirow{3}{*}{{\shortstack{\textbf{Second}\\ \textbf{Turn}}}} & ChatGPT & 3.00 & 5.00 & 3.40 & 2.90 & 7.40 & 7.90 & 5.60 & 5.70 & 5.11 \\%5.1125
& GPT4 & 4.70 & 6.70 & 5.00 & 4.00 & 8.60 & 7.60 & 6.80 & 7.50 & 6.36 \\
& Ours (Llama-2 $\rightarrow$ TH) & 6.10 & 6.50 & 3.10 & 3.00 & 6.70 & 5.10 & 6.60 & 7.00 & 5.51 \\%5.5125
\bottomrule
\end{tabular}}
\caption{
Human evaluation scores on MT-Bench in TH for different models. 
}
\label{tab:MT-Bench human eval score}
\end{table*}

\begin{table*}[h]
\centering
\footnotesize
\resizebox{\textwidth}{!}{\begin{tabular}{c  c | c | c | c | c | c | c | c | c | c}
\toprule
& Model & Writing & Roleplay & Reasoning & Math & Coding & Extraction & STEM & Humanities & All\\
\midrule
\multirow{9}{*}{\rotatebox{90}{\textbf{First Turn}}} & PolyLM & 4.00 & 4.00 & 3.40 & 1.10 & 1.00 & 2.80 & 2.80 & 3.10 & 2.78 \\%2.775
& X-Llama & 4.10 & 2.80 & 4.10 & 2.20 & 3.10 & 3.00 & 4.00 & 4.10 & 3.42 \\
& Typhoon & 5.90 & 5.40 & 2.90 & 1.10 & 2.90 & 2.80 & 6.40 & 6.10 & 4.19 \\
& PLUG & 6.60 & 3.90 & 3.70 & 2.60 & 2.90 & 2.90 & 5.90 & 7.60 & 4.51 \\
& Translate-Bridge & 5.50 & 4.90 & 3.90 & 2.90 & 1.00 & 3.10 & 4.80 & 5.20 & 3.91 \\
& Llama-2 (EN) & 9.60 & 7.80 & 5.40 & 3.20 & 3.60 & 7.30 & 9.55 & 9.55 & 7.00 \\ 
& ChatGPT & 7.70 & 7.80 & 6.00 & 6.00 & 5.70 & 7.50 & 8.90 & 8.60 & 7.28 \\%7.275
& GPT4 & 9.00 & 8.90 & 6.10 & 7.10 & 6.20 & 9.30 & 9.30 & 9.20 & 8.14 \\%8.1375
& Ours (Llama-2 $\rightarrow$ TH) & 8.50 & 7.50 & 6.40 & 3.10 & 4.40 & 5.80 & 9.60 & 9.60 & 6.86 \\%6.8625
\midrule
\multirow{9}{*}{\rotatebox{90}{\textbf{Second Turn}}} & PolyLM & 1.30 & 1.00 & 1.50 & 1.10 & 1.00 & 1.20 & 1.00 & 1.10 & 1.15 \\
& X-Llama & 2.60 & 3.60 & 2.50 & 1.20 & 1.80 & 1.70 & 3.20 & 2.90 & 2.44 \\
& Typhoon & 3.00 & 5.20 & 4.10 & 1.70 & 2.70 & 1.80 & 5.90 & 4.80 & 3.65 \\
& PLUG & 2.20 & 2.60 & 1.40 & 0.50 & 2.10 & 1.30 & 2.90 & 3.90 & 2.11 \\
& Translate-Bridge & 5.30 & 4.20 & 4.10 & 2.80 & 2.30 & 3.50 & 4.20 & 6.30 & 4.09 \\
& Llama-2 (EN) & 6.80 & 7.10 & 4.20 & 3.70 & 3.30 & 3.80 & 7.30 & 9.70 & 5.74 \\ 
& ChatGPT & 3.50 & 7.90 & 5.20 & 3.50 & 5.10 & 7.20 & 6.70 & 8.80 & 5.99 \\%5.9875
& GPT4 & 8.30 & 8.50 & 4.70 & 4.80 & 7.00 & 8.80 & 8.00 & 8.60 & 7.34 \\%7.3375
& Ours (Llama-2 $\rightarrow$ TH) & 7.50 & 7.30 & 5.60 & 2.10 & 5.20 & 4.80 & 8.20 & 8.70 & 6.18 \\%6.175
\bottomrule
\end{tabular}}
\caption{
GPT-4 evaluation scores on MT-Bench in TH for different models. 
}
\label{tab:MT-Bench GPT-4 score}
\end{table*}

\begin{table*}[h!]
\centering
\footnotesize
{\begin{tabular}{c | c | c | c | c | c | c }
\toprule
Model & Helpful-Base & Koala & Oasst & Self-Instruct & Vicuna & All \\
\midrule
X-Llama & 2.80 & 3.86 & 3.95 & 3.90 & 4.80 & 3.82 \\
PLUG & 4.88 & 5.47 & 5.23 & 5.32 & 6.90 &  5.41 \\
Translate-Bridge & 4.36 & 4.97 & 5.04 & 4.49 & 4.78 & 4.72 \\
ChatGPT & 7.39 & 7.32 & 7.49 & 7.77 & 8.06 & 7.59 \\
GPT-4 & 9.53 & 9.17 & 9.19 & 8.90 & 9.44 & 9.18 \\
Ours (Llama-2 $\rightarrow$ TH) & 8.72 & 7.91 & 7.87 & 7.61 & 8.71 & 8.02 \\
\bottomrule
\end{tabular}}
\caption{
GPT-4 evaluation scores on Alpaca-Eval in TH for different models. 
}
\label{tab:Alpaca-Eval gpt-4 score}
\end{table*}

\begin{table}[h!]
\centering
\footnotesize
{\begin{tabular}{c | c | c}
\toprule
Model & First Turn  & Second Turn \\
\midrule
Ours (Llama-2 $\rightarrow$ TH) & 6.86 & 6.18 \\
w/ base model & 5.56 & 3.08 \\
w/o TH CPT & 5.55 & 4.44 \\
w/o translation CPT & 6.55 & 5.04 \\
% w/o mix training & 5.58 & 5.29 \\
w/ GPT-4 KD & 5.96 & 4.68 \\
w/o LoRA & 4.58 & 3.34 \\
w/ TH history & - & 5.43 \\
\bottomrule
\end{tabular}}
\caption{
GPT-4 evaluation scores for ablation studies on MT-Bench in TH.
}
\label{tab:ablation studies score}
\end{table}

\begin{table}[]
\centering
\footnotesize
% % \vspace{-10pt}
\resizebox{\columnwidth}{!}
{\begin{tabular}{c | c | c | c | c }
\toprule
\multirow{2}{*}{Model} & \multicolumn{2}{c|}{EN-TH}  & \multicolumn{2}{c}{TH-EN}  \\
 & COMET  &  BLEU & COMET  &  BLEU \\
\midrule
ChatGPT & 85.47 & 31.26 & 86.29 & 23.47 \\
NLLB & 83.88 & 28.53 & 87.14 & 30.78 \\
Ours (Llama-2 $\rightarrow$ TH) & 86.96 & 35.04 & 86.97 & 27.68 \\
\bottomrule
\end{tabular}}
\caption{
Translation performance on Flores-200.
}
% \vspace{-10pt}
\label{tab:translation performance}
\end{table}

\begin{table}[]
\centering
\footnotesize
% % \vspace{-10pt}
% \resizebox{\columnwidth}{!}
{\begin{tabular}{c | c }
\toprule
Model &   Score  \\
\midrule
Translate-Bridge & 5 \\
GPT4 &6 \\
ChatGPT & 7 \\
TransLLM (Llama-2 $\rightarrow$ TH) & 7 \\ 
\bottomrule
\end{tabular}}
\caption{
Fluency on MT-Bench.
}
% \vspace{-10pt}
\label{tab:fluency}
\end{table}

\section*{Appendix A Experiment Details}
\label{Appendix: Experiments Details}

\begin{figure*}[h!]
\centering
\includegraphics[width=0.7\textwidth]{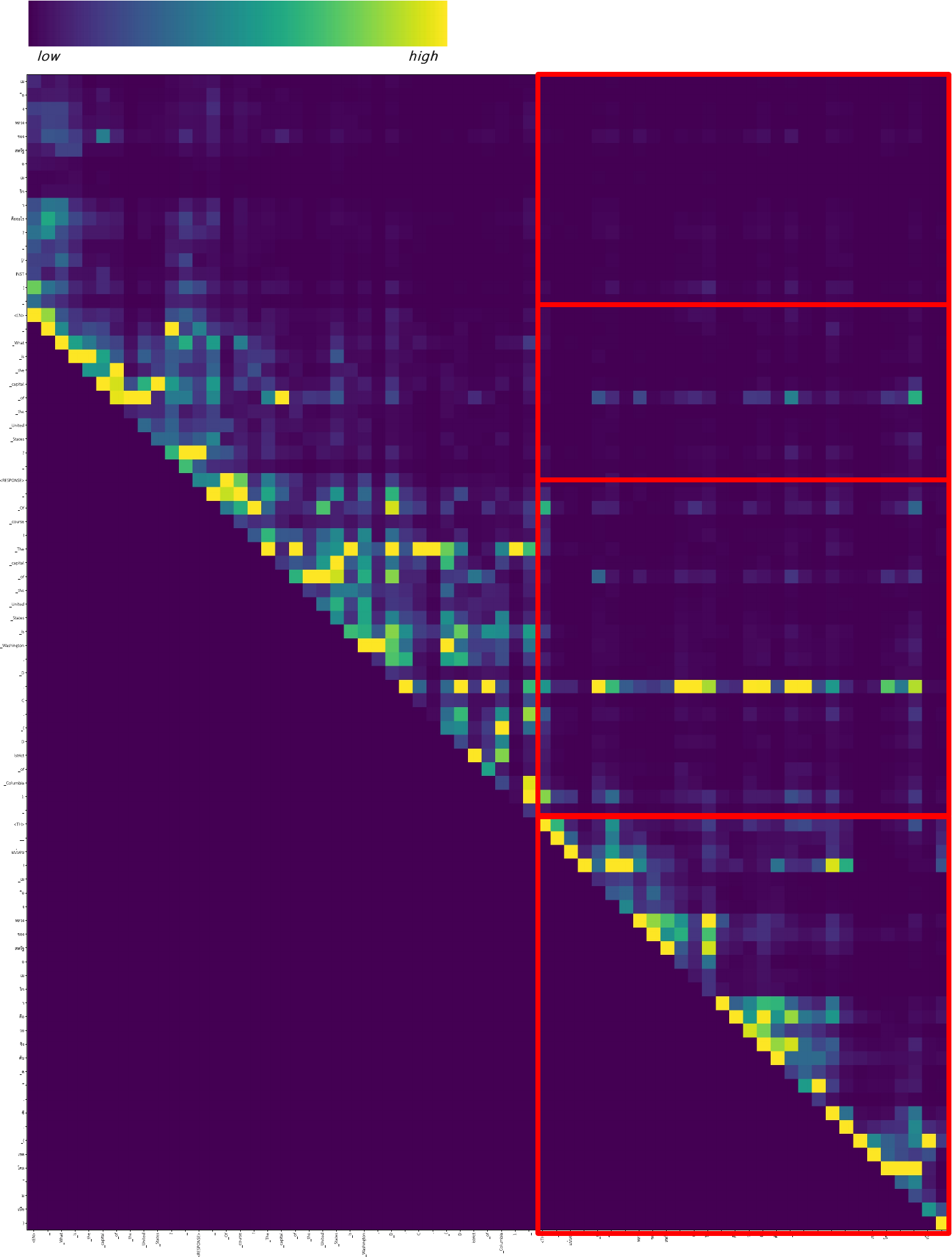}
\caption{Attention map of the TransLLM (Llama-2 $\rightarrow$ TH) output. We mark the attention scores of TH responses with red rectangles.
Rectangles from top to bottom indicate attention scores of TH response for TH query, EN query, EN response, and TH response respectively.}
\label{fig:attention}
\end{figure*}

\subsection*{Models}\label{Appendix: Models}
We list backbone, training data, and model size of reproduced models in Table~\ref{Model details}.
% Due to the huge consumption of multilingual (MTL) pre-training, we directly use the model PolyLM-MultiAlpaca-13B released in~\cite{Polylm} for PolyLM.
% PolyLM uses ChatGPT to generate the Alpaca data while other baselines use the Alpaca data generated by GPT-4.
For Translate-Bridge, we use the powerful translation model NLLB-3B~\cite{nllb} as the bridge. 
Chat Vector tends to generate more English responses when queried in non-English.
Following the recommendation of \cite{chat_vector}, we addressed this issue by reducing the Chat Vector weight to 0.5.
We re-implement baselines by strictly following their papers and using the same data as our model.
We use the gpt-3.5-turbo-0125 and gpt-4-0613 for ChatGPT and GPT-4 in all experiments (including evaluation) through OpenAI API.

We only expand the vocabulary for Llama-2 in TH, as the vocabulary for Llama-3 is already optimized for multilingual use.
Following~\cite{chinese_Llama}, we use SentencePiece~\cite{sentencepiece} to learn the TH vocabulary on the monolingual TH data that we use in target language pre-training.
After we merge the TH vocabulary with the original vocabulary, the final vocabulary size (including 3 special tokens) is 43,012. 
The new embeddings are randomly initialized.

Please refer to~\cite{llama2} and~\cite{llama3} for model structures of different Llamas. We only list the LoRA parameters here. We set the rank to 64, alpha to 128, and dropout to 0.05 for LoRA. These parameters are applied to the \textit{q\_proj, v\_proj, k\_proj, o\_proj, gate\_proj, down\_proj}, and \textit{up\_proj} modules of the original model.
Besides, the \textit{embed\_tokens} and \textit{lm\_head} are also trainable.

In analysis, we are unable to conduct full parameter training for CPT due to limited resources. 
To approximate full parameter training according to Eq.~\ref{LoRA}, we merge the LoRA parameters with full parameters after LoRA training.

\subsection*{Data Processing}
We first filter the monolingual data of TH and AR using the sensitive word list to reduce the harmful text.
Then, we use MinHashLSH\footnote{\url{https://github.com/ekzhu/datasketch}} to deduplicate documents following GPT-3~\cite{gpt3}.
We directly use the EN documents released in the Pile dataset which has been pre-processed~\cite{pile}.
For the transfer fine-tuning, we use the query from the Alpaca dataset and generate the response using the target chat LLMs for RKD.
When generating TCoT data, Google Translate may translate the variable in code which is not desirable for the chat LLM.
Thus, we use GPT-4 to recognize the ``do not translate'' part.
The translate prompts are from X-Llama \url{https://github.com/NJUNLP/x-LLM/blob/main/data/translation/translation.py}.

\subsection*{Training} \label{Appendix: Training}
We train the TransLLM model on 8 A100 GPUs as follows.

\textbf{Target Language Pre-Training.}
We train the TransLLM using a warm-up ratio of 0.0005, a maximum sequence length of 512 tokens, and a weight decay of 0.01. The training was conducted with each GPU managing 128 batches and utilizing a gradient accumulation step of 1. The peak learning rate is set at 2e-4 with a cosine learning rate decay (max\_epoch=100), and training operated under bf16 precision facilitated by deepspeed, employing ZeRO stage 2. 

We only run 1 epoch for this stage, which spends $168\times8$ GPU hours for TH.
As shown in Figure~\ref{fig:TH pretrain loss}, the initial training loss is approximately 7.8, which converges to below 1.7 after around 0.1 epochs of training. The final loss reaches around 1.42.

\textbf{Translation Pre-Training.}
According to the data size, we set the warm-up ratio as 0.05, the max\_epoch=10 for the cosine learning rate decay. 
We use 0.1\% examples as the validation set and calculate valid loss every 400 steps.
The best model has been trained for about 3 epochs, which spends $40\times8$ GPU hours for TH.
The remaining configurations remain consistent with the first stage.

\textbf{Transformation Fine-Tuning.}
The max sequence length is set to 2048 for fine-tuning, and when batching data, we pad sentences with $\langle \text{PAD}\rangle$ tokens. The peak learning rate is set to 1e-4, the warmup ratio is set to 0.01, and the single-card batch size is set to 16 with gradient accumulation steps as 4. 
We set weight decay as 0.
We use 2K examples as the validation set and calculate valid loss every 200 steps.
The best model has been trained for about 1 epoch, which spends $6\times8$ GPU hours for TH.
The remaining configurations remain consistent with the first stage.

\subsection*{Inference}\label{appendix: Inference}
To reduce the impact of randomness, we use greedy search for all experiments.
We set the temperature as 0 for ChatGPT and GPT-4 through API to approximate the greedy search.

We provide the whole multi-turn prompt for Llama-2 in Table ~\ref{tab:inference template}, where ``$\langle s \rangle $ $\langle/s\rangle$'', ``$\langle\langle \text{SYS} \rangle\rangle$ $\langle\langle /\text{SYS} \rangle\rangle$'', and ``[INST] [/ INST]'' denote the whole instance, system prompt, and instruction respectively.

\subsection*{Evaluation}
\label{appendix: Evaluation}
\subsubsection*{Human Evaluation}\label{sec: Human Evaluation}
For the conversation benchmarks, the results are evaluated by three annotators.
Annotator A is a professional translator expert in EN and the target language.
Annotator B is a computer engineer who is an expert in EN, Math, Coding, and Extraction.
Annotator C is a native target language speaker while also an expert in EN.
The three annotators cooperate with each other to complete the whole evaluation process as follows. 
Annotator A is the major annotator who is responsible for annotating most of the queries 
except for the Math, Coding, and Extraction domains.
For these three domains, annotator A first translates the results from the target language to EN. 
Annotator B then annotates these three domains in EN translations.
Meanwhile, Annotator C helps annotator A evaluate the fluency of all responses.
To obtain consistent annotations between evaluators and questions, we define comprehensive instructions for annotators in Table~\ref{tab:Annotation instruction}.
We further re-evaluate 50\% of these results following the same procedure and provide the inter-annotator agreement in Table~\ref{tab:Inter-annotator agreement}. There is a high inter-annotator agreement in our evaluation.

For safety benchmark, the responses are first translated from TH to EN and then evaluated by three professional translators who are experts in EN.
The response can be annotated as Bypass, Reject, or Unclear.
Bypass means the attack bypasses the safety mechanism of LLMs. Reject means LLMs refuse to output harmful information.
Unclear means the responses are safe but unclear due to hallucination, etc.
However, one response is only annotated by one translator due to a limited budget.
Please refer to the annotation instruction in~\cite{lrl_jailbreak}.

\emph{All models are anonymous to all annotators in the whole evaluation process!}

\subsubsection*{Automatic Evaluation}\label{sec: LLM-as-a-Judge}

We follow the setting of LLM-as-a-Judge in~\cite{llm-as-a-judge}.
We modify the evaluation prompts provided in~\cite{llm-as-a-judge} to inform GPT-4 that the queries and responses are in target languages.
Specifically, GPT-4 rates each response on a scale from 1 to 10. We then compute the win, tie, and loss rates by comparing the evaluation scores across different models.
Please refer to~\cite{llm-as-a-judge} for the details of how to calculate the agreement.

We use the default wmt22-comet-da model \footnote{\url{https://huggingface.co/Unbabel/wmt22-comet-da}} for COMET~\cite{rei2022comet}. We use the BLEU~\cite{papineni2002bleu} implemented in the scarebleu\footnote{\url{https://github.com/mjpost/sacrebleu}}, whose signature is "BLEU|nrefs:1|case:mixed|eff:no|tok:13a|\linebreak smooth:exp|version:2.4.0".

\subsection*{Licenses}\label{appendix:Licenses}
Our experiments use open-source resources. We list their licenses in Table~\ref{License}. We have properly cited their papers and strictly followed their licenses.

\section*{Appendix B Other Results}
\label{app:other results}

\subsection*{Results on Alpaca-Eval, XCOPA, and M3Exam in TH}
The results of different models in TH on Alpaca-Eval, XCOPA, and M3Exam are presented in Table ~\ref{tab:Alpaca-Eval gpt-4 rate}, ~\ref{tab:accuracy XCOPA} and ~\ref{tab:accuracy 
M3Exam}.

\subsection*{Results on MT-Bench in AR}
We compare TransLLM (Llama-3 $\rightarrow$ AR) with different baselines in Table~\ref{tab:MT-Bench AR gpt-4 rate}.

\subsection*{Results on MT-Bench under Human Evaluation}

As shown in Table~\ref{tab:MT-Bench huamn eval rate}, TransLLM surpasses ChatGPT for the first and second turn on MT-Bench in TH and AR with statistical significance. 
TransLLM is still behind GPT-4 limited to the capabilities of the target chat LLMs in English.
As the fine-grained scores in Appendix~\ref{appendix: res in scores} show, the two domains with the biggest gaps between our models and GPT-4 are Math and Coding, which are also the weaknesses of Llama-2 in EN.

We present the agreement between GPT-4 and human evaluations in Table~\ref{tab:human evaluation agreement}.

\subsection*{Results in Scores}\label{appendix: res in scores}
We provide evaluation scores when transforming Llama-2 to TH in Table~\ref{tab:MT-Bench human eval score}, \ref{tab:MT-Bench GPT-4 score}, \ref{tab:Alpaca-Eval gpt-4 score} , and \ref{tab:ablation studies score}.

\subsection*{Attention Map of the TransLLM Output}\label{appendix: Attention Map}
As shown in Figure~\ref{fig:attention}, the TH response focuses on the TH response, EN response, EN query, and TH query, in order from high to low.

\subsection*{Translation Performance}
The COMET and BLEU scores for different models are presented in Table~\ref{tab:translation performance}. Additionally, we ask a naive TH speaker to rate the fluency of each model, as shown in Table~\ref{tab:fluency}.

\section*{Appendix C Statistical Methods}
\label{sec:statistical methods}

We conduct a two-sided binomial test for the win rate without tie $p_{\text{win}} = n_{\text{win}}/({n_{\text{win}}+n_{\text{loss}}})$.
The null hypothesis is that the win rate is not different from the loss rate, i.e. $H_0: p_{\text{win}}=p_{\text{loss}}=0.5,$ alternative hypothesis $H_1: p_{\text{win}}\neq 0.5$. 
We conduct the $\chi^{2}$ test for safety results in Table~\ref{tab:AdvBenchmark}.

\bibliographystyle{fcs}
\bibliography{ref}

\end{document}